\documentclass[journal]{IEEEtran}
\usepackage{amsmath,amsfonts}
\usepackage{algpseudocode}
\usepackage{algorithmicx,algorithm}
\usepackage{hyperref}
\usepackage{array}
\usepackage[caption=false,font=normalsize,labelfont=sf,textfont=sf]{subfig}
\usepackage{textcomp}
\usepackage{stfloats}
\usepackage{url}
\usepackage{verbatim}
\usepackage{graphicx}
\usepackage{cite}

\usepackage[square,sort&compress,numbers]{natbib}
\usepackage{amsmath}
\usepackage{amsfonts} 
\usepackage{amssymb} 
\usepackage{multirow}
\usepackage{threeparttable}
\usepackage{booktabs}
\usepackage{bbm}
\usepackage{bbding}
\usepackage{pifont}

\usepackage{bbm, dsfont}
\usepackage{ifsym}

\graphicspath{{figs/}}
\graphicspath{{bio/}}

\usepackage{color, xcolor}

\begin{document}

\title{Semantic-CC: Boosting Remote Sensing Image Change Captioning via  Foundational Knowledge and Semantic Guidance
}

\author{
Yongshuo~Zhu,~Lu~Li$^\star$,~Keyan~Chen,~Chenyang~Liu,~Fugen~Zhou~and~Zhenwei~Shi
\\
\vspace{4pt}
Beihang University
}

\maketitle

\setlength{\abovedisplayskip}{6pt}
\setlength{\belowdisplayskip}{2pt}

\begin{abstract}

Remote sensing image change captioning (RSICC) aims to articulate the changes in objects of interest within bi-temporal remote sensing images using natural language. Given the limitations of current RSICC methods in expressing general features across multi-temporal and spatial scenarios, and their deficiency in providing granular, robust, and precise change descriptions, we introduce a novel change captioning (CC) method based on the foundational knowledge and semantic guidance, which we term Semantic-CC. Semantic-CC alleviates the dependency of high-generalization algorithms on extensive annotations by harnessing the latent knowledge of foundation models, and it generates more comprehensive and accurate change descriptions guided by pixel-level semantics from change detection (CD). Specifically, we propose a bi-temporal SAM-based encoder for dual-image feature extraction; a multi-task semantic aggregation neck for facilitating information interaction between heterogeneous tasks; a straightforward multi-scale change detection decoder to provide pixel-level semantic guidance; and a change caption decoder based on the large language model (LLM) to generate change description sentences.
Moreover, to ensure the stability of the joint training of CD and CC, we propose a three-stage training strategy that supervises different tasks at various stages. We validate the proposed method on the LEVIR-CC and LEVIR-CD datasets. The experimental results corroborate the complementarity of CD and CC, demonstrating that Semantic-CC can generate more accurate change descriptions and achieve optimal performance across both tasks.

\end{abstract}

\begin{IEEEkeywords}
 Remote sensing image, change captioning, foundation model, multi-task learning
\end{IEEEkeywords}

\IEEEpeerreviewmaketitle
\section{Introduction}

\IEEEPARstart{R}emote sensing image change interpretation technology has emerged as a pivotal tool across various domains, including environmental monitoring, urban planning, and disaster management \cite{du2014remote, shafique2022deep, chen2023time, chen2022resolution, zhang2024bifa, chen2023continuous}. Remote sensing image change captioning (RSICC) constitutes a critical aspect of this technology, focusing on describing the changes between temporally disparate remote sensing images through natural language \cite{liu2022remote, hoxha2022change, liu2024rscama}. The advent of specialized datasets and the refinement of vision-language models have catalyzed substantial advancements in deep-learning-driven RSICC methods \cite{liu2024rscama, liu2023decoupling}.

Despite significant progress in the field of RSICC, current methodologies still struggle with performance in environments with architectural changes, varying lighting conditions, and low-contrast scenarios \cite{chang2023changes, shen2020remote}. This limitation is largely due to an over-reliance on deep semantic information for text generation, which tends to neglect the finer details of the images \cite{hoxha2022change}. Occasionally, these methods may become ensnared by semantic noise unrelated to the change understanding task, such as shifts in the angle and intensity of lighting \cite{chen2023time}.
These factors hinder the accurate discrimination of meaningful changes. To address this issue, Liu et al. \cite{liu2023pixel} have introduced a novel approach that incorporates multi-task learning, where the remote sensing image change detection (RSICD) task is leveraged as a supplementary component. By harnessing pixel-level information and feeding it into the RSICC model through a shared image encoder, this strategy significantly enhances the model's ability to recognize and express effective changes, thus improving the overall accuracy.

Indeed, utilizing RSICD as an auxiliary task to enhance the semantic expression of RSICC models presents a promising approach. This is primarily due to the fact that RSICD, which encompasses the identification and precise localization of pixel-level changes, provides crucial insights that are pivotal in formulating detailed and accurate descriptions of the differences between bi-temporal images. Nonetheless, integrating these two tasks within a single framework poses significant challenges \cite{wang2024rsbuilding, liu2024change}.
Firstly, the scarcity of annotated datasets, which incorporate both pixel-level and semantic-level annotations, frequently leads to the overfitting of deep learning models. Secondly, most existing models are custom-made for a specific task. When faced with multiple tasks, these models grapple with bridging the semantic chasm between high-level semantic information and detailed pixel-level specifics. These challenges complicate the integration of homogeneous features from different tasks within a unified framework \cite{bai2023deep, liu2022remote}.

Recently, vision-language foundation models have exhibited remarkable generalization capabilities, demonstrating superior performance in domain-specific transfer tasks \cite{radford2021learning, alayrac2022flamingo,li2023blip,liu2024visual,zhu2023minigpt, chen2024rsprompter}. By harnessing vast amounts of data for large-scale training, these models can learn generalized representations, thereby efficiently transferring and sharing general knowledge across various specific downstream tasks \cite{kirillov2023segment, chen2024internvl, chen2024rsmamba}. The introduction of foundation models' latent knowledge can ensure performance while liberating from the dependence on extensive annotations. Moreover, large-scale vision-language models possess an enhanced potential to bridge the semantic-level and pixel-level information, thereby augmenting the overall comprehension and interpretation of visual inputs \cite{wang2024mtp, lu2023viewpoint}.

In this paper, we introduce Semantic-CC, a novel remote-sensing image change captioning method that integrates foundational knowledge and semantic guidance. This method alleviates the dependency of high-generalization algorithms on extensive annotations by leveraging the latent knowledge inherent in foundation models. Moreover, it enhances the description of changes, making them more granular, robust, and precise through pixel-level semantic guidance from change detection. Specifically, Semantic-CC consists of four components: a bi-temporal SAM-based encoder for dual-image feature extraction; a multi-task semantic aggregation neck for facilitating information interaction across heterogeneous tasks; a change detection decoder for providing pixel-level semantic guidance; and a change caption decoder for generating descriptive sentences of changes.

\vspace{7pt}
The primary contributions of this paper can be summarized as follows:

i) We introduce Semantic-CC, a CC method that integrates foundational knowledge and semantic guidance. This method can be trained with minimal annotations to achieve high usability and provide more granular and accurate sentence descriptions.

ii) Specifically, we propose a bi-temporal SAM-based encoder, which is constructed on the latent knowledge of the SAM foundation model and incorporates a bi-temporal change semantic filter to integrate dual-temporal information into the foundational encoder. We also propose a multi-task semantic aggregation neck for intra- and inter-task feature attention; a straightforward multi-scale CD decoder to guide the finer granular semantic expression of CC; and a CC decoding head based on Vicuna, which includes a change semantic feature enhancer that generates bi-temporal differential features for the expression of change semantics.

iii) To ensure the smooth convergence of CD and CC and to prevent negative transfer in multi-task learning, we design a three-stage training strategy to supervise the training of different tasks at different stages.

iv) We validate our method through experiments on the LEVIR-CD and LEVIR-CC datasets. The experimental results demonstrate the complementarity of the semantics of CC and CD and suggest that Semantic-CC can generate high-precision CD masks and CC descriptions, thereby achieving superior performance in both tasks.

\section{Related Works}

\subsection{Remote Sensing Image Change Detection}

Remote sensing image change detection is designed to analyze pixel-level changes between temporally distinct images, which are represented by a binary change mask, where zero signifies non-changes and one denotes changes \cite{zhang2024cdmamba, zhang2024bifa, chen2023time}. Recent methods predominantly concentrate on bi-temporal remote sensing image analysis utilizing deep-learning techniques \cite{chen2020dasnet, chen2023time}.

Most deep learning methods for change detection employ architectures based on convolution or Transformer to facilitate supervised learning from annotations, with the primary objective of discerning ``effective changes". These endeavors are committed to augmenting the models' feature representation and discrimination capabilities, which specifically encompass various bi-temporal feature fusion strategies \cite{li2022remote,zheng2022changemask,lei2021difference, zhang2024bifa}, multi-scale feature aggregation methods \cite{rahman2018siamese,hou2019w,daudt2018fully, zhang2024cdmamba}, and diverse spatiotemporal attention modules \cite{chen2020dasnet,jiang2020pga, chen2020spatial,chen2021remote}. For instance, Li et al. \cite{li2022remote} proposed a temporal feature interaction module to facilitate interaction between multi-level bi-temporal features of convolutional neural networks (CNN) and to obtain multi-level differential features. Peng et al. \cite{peng2020optical} introduced a dense attention method that utilizes high-level features with category information to guide the selection of low-level features.

While supervised learning can yield quite satisfactory results and has become mainstream in change detection, the requirement for a large volume of labeled data has emerged as a bottleneck for its further advancement. Some researchers have started to incorporate unsupervised learning methods and the latent knowledge of foundation models \cite{li2024new}. For example, Saha et al. \cite{saha2019unsupervised} proposed a deep change vector analysis method that combines change vectors with spatial context information extracted by pre-trained CNN to identify the changed pixels. Ren et al. \cite{ren2020unsupervised} proposed a change detection model based on graph convolutional networks and metric learning, which employs a multi-scale dynamic graph convolutional network module to capture contextual information and extract spatial-spectral features, and combines spatial-spectral feature analysis with metric learning to generate reliable pseudo labels for unsupervised training.

With the evolution of large foundation models such as SAM \cite{kirillov2023segment}, some researchers have also begun to explore the downstream applications of foundation models in the field of change detection in remote sensing images. Based on the SAM, Chen et al. \cite{chen2023time} designed a time-traveling gate module to fuse the information of pre-temporal and post-temporal images, thereby enhancing the model's sensitivity to effective changes while ignoring ineffective ones. The development of foundation models has brought new possibilities to the RSICD.

The evolution of RSICD technology has led to significant achievements in the field. However, the progress of RSICD algorithms continues to encounter several challenges. Firstly, the majority of existing RSICD datasets are characterized by low quality and limited volume. Consequently, the development of high-quality, enriched RSICD datasets has become a focal point of research. Secondly, the extensive use of high-resolution remote sensing images has resulted in an increased diversity of land cover types and a heightened complexity of change features. This trend necessitates the development of more effective and accurate detection methods. Lastly, the diverse application scenarios of RSICD have imposed new demands on efficiency.

\subsection{ Remote Sensing Image Change Captioning}

Remote sensing image change captioning represents a burgeoning multimodal task that integrates remote sensing image processing with natural language generation \cite{liu2024change}. Currently, deep learning methods are predominantly employed to navigate the complexities inherent in this task. The foundational architecture for RSICC mirrors that utilized in the image captioning task, with both leveraging an encoder-decoder framework that incorporates CNNs or Transformers \cite{Liu2022Remote2}. However, RSICC necessitates a concentrated focus on alterations within multi-temporal images, in contrast to the image captioning task which merely require an understanding of a singular image. The encoder in change captioning is tasked with extracting changed semantic features from bi-temporal images, while the decoder is responsible for translating the change representation into natural language, employing technologies such as recurrent neural networks (RNN) or Transformers.

Pioneering work by Chouaf and Hoxha et al. \cite{chouaf2021captioning,hoxha2022change} involved the use of pre-trained CNNs as encoders and attempted to employ RNNs and support vector machines (SVM) as decoders to generate text descriptions. To further propel research in change captioning, Liu et al. \cite{liu2022remote} proposed a large-scale change captioning dataset, LEVIR-CC, and benchmarked several methods. They also enhanced the perception of objects of varying sizes in change captioning through a progressive scale perception network with a change perception layer and a scale perception enhancement module \cite{liu2023progressive}. Chang et al. \cite{chang2023changes} proposed an attention network that includes a hierarchical self-attention module and a residual unit for identifying attributes related to changes and constructing semantic change embeddings. These methods, based on the traditional encoding-decoding structure, represent a classic solution for RSICC tasks. Recently, with the evolution of vision-language models, high-performance RSICC methods based on these foundation models have also begun to be explored. Liu et al. \cite{liu2023decoupling} decoupled the task of RSICC into a binary classification task to determine the presence of changes and a text generation task, using the pre-trained CLIP model and a large language model to enhance the performance of change captioning. However, the exploration and application of vision-language models in the RSICC field are still in their infancy, and further research into RSICC methods based on large vision-language models is a promising direction.

Compared to RSICD, RSICC is a relatively nascent research direction. Over recent years, there has been a significant accumulation of technology and data in the RSICC field. Coupled with the advancement of cross-modal deep learning techniques, it has propelled RSICC to the forefront of research interest. Nevertheless, current RSICC models face challenges in generating precise, robust, and comprehensive language descriptions. This is particularly evident in the description of small targets and spatial locations. Therefore, RSICC necessitates further exploration and research.

\subsection{Multi-Task Learning}

Multi-task learning (MTL) is a strategy that seeks to harness a singular model to manage multiple tasks concurrently. This approach has the potential to augment data utilization efficiency and mitigate overfitting through the learning of shared representations \cite{zhang2021survey, vandenhende2021multi}. Furthermore, it can bolster the robustness of the model by capitalizing on the complementary information derived from multiple tasks. This multi-pronged approach not only streamlines the learning process but also enhances the overall performance and reliability of the model \cite{wang2024rsbuilding}.

Owing to its capacity to manage multiple tasks within a unified architecture, MTL has gained considerable traction in natural language processing (NLP) and computer vision (CV). In the NLP domain, Liu et al. \cite{liu2016recurrent} proposed three distinct shared information mechanisms based on RNNs to model specific tasks, yielding commendable results in four text classification tasks. Similarly, Luong et al. \cite{luong2015multi} explored three MTL modes for sequence-to-sequence models. In CV, several methodologies also concurrently learn multiple visual tasks \cite{liu2019end,standley2020tasks,zamir2020robust, chen2023ovarnet}. For instance, Chen et al. \cite{chen2021pre} proposed a large-scale MTL network with Transformers, aimed at resolving various visual tasks by fine-tuning pre-trained models. Mohamed et al. \cite{mohamed2021spatio} suggested joint learning of object detection and semantic segmentation tasks. These methodologies underscore the superiority of MTL paradigms.

Beyond utilizing a single model to manage multiple sub-tasks, the construction of auxiliary tasks within the MTL framework to enhance the performance of the primary task has also become a prevalent paradigm \cite{chen2021building}. Hosseinzadeh et al. \cite{hosseinzadeh2021image} devised an auxiliary network to characterize the changes between two images, thereby enabling the primary model to generate more detailed and accurate captions. Xu et al. \cite{xu2023auxiliary} proposed an additional adaptive Transformer to accomplish the task of human motion prediction. Ak et al. \cite{ak2022learning} integrated the text-based image manipulation (TIM) task with the CC task, thereby not only enhance the training of the TIM module but also leveraging the TIM module as additional supervision for CC training.

MTL involves sharing parameters between tasks and learning a common representation of multiple tasks to achieve joint optimization between different tasks. However, MTL also presents certain challenges, such as conflicts and imbalances between tasks. Addressing these challenges remains a focal point of MTL research.

\subsection{ Foundation Model in Remote Sensing}

Large language models (LLM) have significantly advanced the progression of general artificial intelligence (AGI) and have extended their influence into the realm of remote sensing image intelligence analysis. In an effort to further enhance this field, researchers have initiated the development of large-scale multimodal vision-language models (VLM) for remote sensing images, aiming for more comprehensive, generalized, and stable interpretive approaches \cite{chen2024rsprompter, kuckreja2024geochat, hu2023rsgpt}. At present, foundation models in the remote sensing domain primarily bifurcate into two research trajectories: pre-training on extensive remote sensing image data, and fine-tuning existing foundation models on a smaller scale but with high-quality data.

In the context of pre-training, several notable works have been introduced. Sun et al. \cite{sun2022ringmo} proposed a foundational remote sensing model framework, RingMo, which employs 2 million remote sensing image data and generative self-supervised learning for pre-training. Yao et al. \cite{yao2023ringmo} suggested a spatiotemporal prediction foundation model predicated on spatiotemporal evolution deblurring. Wang et al. \cite{wang2022advancing} trained models including CNNs and Vision Transformers (ViT) based on the MillionAID dataset and further explored the influence of different pre-training regimes on a range of downstream tasks. Large-scale pre-training necessitates substantial computing resources, thus efficient fine-tuning of large models under resource-constrained conditions warrants further investigation.

Hu et al. \cite{hu2023rsgpt} introduced a high-quality remote sensing image captioning dataset, RSICap, and devised a scheme for fine-tuning InstructBLIP \cite{liu2024improved}, proposing the remote sensing generative pre-training model (RSGPT) to swiftly align the visual features of remote sensing images with the semantic features of LLMs. Inspired by prompt learning, Chen et al. \cite{chen2024rsprompter} designed an automated instance segmentation method based on the SAM foundation model, capable of generating suitable prompts for SAM's input, thereby enabling it to yield semantically distinguishable segmentation results for remote sensing images. In TTP \cite{chen2023time}, Chen et al. further expanded their work, designing a time-traveling gate module based on the SAM to incorporate the pre- and post-temporal information of bi-temporal remote sensing images, thereby enhancing the foundation model's sensitivity to effective changes in change detection tasks, while disregarding irrelevant information.

Models trained on extensive data have amassed a wealth of general knowledge, facilitating the possibility of cross-domain knowledge transfer and sharing. Large-scale pre-training and efficient fine-tuning for specific downstream tasks represent the future directions of the remote sensing image processing field and hold considerable potential for processing remote sensing images with variable spatiotemporal and scale characteristics.

\begin{figure*}[htbp]
    \centering
    \includegraphics[width=\linewidth]{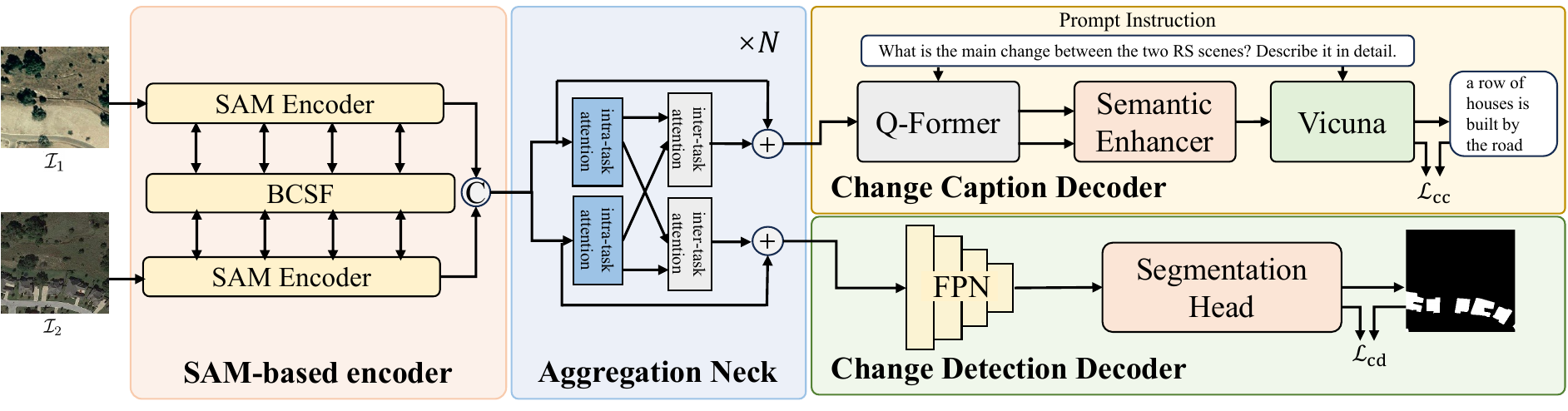}
    \caption{The architecture of the Semantic-CC consists of four main components: a bi-temporal SAM-based encoder, a multi-task semantic aggregation neck, a change detection decoder, and a change caption decoder.}
    \label{fig:model}
\end{figure*}

\section{Methodology}

\subsection{Architecture}

In this paper, we present Semantic-CC, an innovative method for remote sensing image change captioning, which is informed by the semantic guidance of change detection and the prior knowledge from the foundation model. Semantic-CC employs a bi-temporal image input and dual-branch decoding architecture, thereby facilitating a more comprehensive and precise interpretation of bi-temporal remote sensing images. The overarching structure is depicted in Fig. \ref{fig:model} and primarily comprises four key components: a bi-temporal SAM-based encoder for the image feature extraction; a multi-task semantic aggregation neck for information interaction between heterogeneous tasks; a change detection decoder for supplying pixel-level semantic guidance and predicting change segmentation masks; and a change caption decoder for generating change description sentences. The entire process can be described by the following formula,
\begin{align}
    \begin{split}
        F_1, F_2 &= \Phi_{\text{sam-enc}} ( \mathcal{I}_1, \mathcal{I}_2) \\
        F_\text{cd}, F_\text{cc} &= \Phi_{\text{agg-neck}} ( F_1, F_2) \\
        M &= \Phi_{\text{cd-decode}} ( F_\text{cd} ) \\ 
        T &= \Phi_{\text{cc-decode}} ( F_\text{cc}, P ) \\
    \end{split}
\end{align}
where the pre- and post-temporal images ($\mathcal{I}_1 \in \mathbb{R}^{H \times W \times 3}$ and $\mathcal{I}_2 \in \mathbb{R}^{H \times W \times 3}$) are encoded by $\Phi_{\text{sam-enc}}$ to derive bitemporal features ($F_1 \in \mathbb{R}^{\frac{H}{16} \times \frac{W}{16} \times c}$ and $F_2 \in \mathbb{R}^{\frac{H}{16} \times \frac{W}{16} \times c}$). These bitemporal features are subsequently inputted into $\Phi_{\text{agg-neck}}$ to facilitate task semantic interaction, resulting in the generation of task-sensitive features, $F_\text{cd} \in \mathbb{R}^{\frac{H}{16} \times \frac{W}{16} \times c}$ and $ F_\text{cc} \in \mathbb{R}^{\frac{H}{16} \times \frac{W}{16} \times c}$, to produce the change segmentation map ($M \in \mathbb{R}^{H \times W}$), and the change caption ($T \in \mathcal{C}^{N}$) ($\mathcal{C}$ is the captioning vocabulary, and $N$ is the caption length), respectively. $\Phi_{\text{cd-decode}}$ and $\Phi_{\text{cc-decode}}$ are the decoders for the change detection and the change captioning respectively. $P$ symbolizes the prompt instruction of change captioning.

\subsection{Bi-temporal SAM-based Encoder}

In change detection and captioning within bitemporal images, the primary objective is to discern exact changes while effectively mitigating the influence of extraneous variables such as lighting conditions, atmospheric effects, and rectification discrepancies. Foundation models have demonstrated considerable efficacy in capturing salient features within intricate scenes. However, there exists a notable limitation in the existing encoders of these foundation models, which excel in extracting features from individual images but exhibit a relative deficiency in processing dual images. This shortcoming is particularly pronounced when it comes to uncovering the differential characteristics inherent to bi-temporal images.

\begin{figure}[!htbp]
    \centering
    \includegraphics[width=0.8\linewidth]{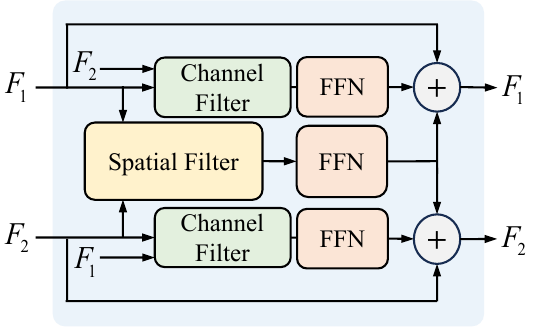}
    \caption{The structure of the bi-temporal change semantic filter (BCSF).}
    \label{fig:BCSF}
\end{figure}

To enhance the capabilities of the foundation model, thereby enabling them to discern meaningful changes between dual images and to circumvent semantic ambiguities, we have integrated a bi-temporal change semantic filter (BCSF) into the encoder of the SAM foundation model, as depicted in Fig. \ref{fig:BCSF}. The BCSF operates on both spatial and channel perspectives, thereby facilitating the transmission and filtration of change semantic features. Specifically, four instances of the BCSF have been strategically positioned after the four global attention layers within the SAM ViT encoder \cite{dosovitskiy2020image}. The functionality of the BCSF can be encapsulated by the following mathematical formulation,
\begin{align}
    \begin{split}
        F_1^{\text{sf}} &= \Phi_{\text{ffn}}^{\text{sf}} (\Phi_{\text{sf}} (F_1, F_2) ) \\
        F_2^{\text{sf}} &= \Phi_{\text{ffn}}^{\text{sf}} ( \Phi_{\text{sf}} (F_2, F_1)) \\
        F_1^{\text{cf}} &= \Phi_{\text{ffn}}^{\text{cf-1}} (\Phi_{\text{cf-1}} (F_1, F_2) )\\
        F_2^{\text{cf}} &= \Phi_{\text{ffn}}^{\text{cf-2}} (\Phi_{\text{cf-2}} (F_2, F_1) )\\
        F_k &= F_k + F_k^{\text{sf}} + F_k^{\text{cf}}, k \in \{1, 2\}\\
    \end{split}
\end{align}
where $\Phi^{\text{sf}}$ and $\Phi^{\text{cf-}k}$ are used to represent the spatial and channel bi-temporal change semantic filters, respectively. $\Phi_{\text{ffn}}^{\text{sf}}$, $\Phi_{\text{ffn}}^{\text{cf-1}}$, and $\Phi_{\text{ffn}}^{\text{cf-2}}$ refer to feed-forward networks (FFN) that are parameterized differently. $F_k^{\text{sf}}$ and $F_k^{\text{cf}}$ signify the image features that result from spatial and channel filtering, respectively. $F_k \in \mathbb{R}^{\frac{H}{16} \times \frac{W}{16} \times c}$ denotes the filter's output with bi-temporal interaction, with $k$ representing the different time points.

Considering the spatial feature consistency inherent in bi-temporal remote sensing images, we have adopted a shared parameter spatial filter, $\Phi_{\text{sf}}$. However, it is recognized that different temporal phases may display unique characteristics within the channel dimension. We apply distinct channel filters for different time phases, $\Phi_{\text{cf-1}}$ and $\Phi_{\text{cf-2}}$. The spatial and channel filters are mathematically defined by the following equations,
\begin{align}
    \begin{split}
        \Phi_{\text{sf}/\text{cf}}(x, y) &= x \otimes \sigma( \Phi_{\text{proj}} (\Phi_{\text{cat}} (x, y) ))\\
    \end{split}
\end{align}
where we leverage a straightforward methodology for the spatial/channel feature integration of the bi-temporal images.
In the spatial filter, $x$ symbolizes the feature subjected to filtering, and $y$ refers to another temporal counterpart that informs the filtering criteria. $\Phi_{\text{cat}}$ denotes feature concatenation along the channel dimension. The transformation of channel numbers from $2 \times c$ to 1 is achieved through a linear layer, $\Phi_{\text{proj}}$. $\sigma$ is the sigmoid activation function. $\otimes$ signifies the pixel-wise multiplication. The channel filter mirrors the aforementioned spatial filtering process, with the key distinction that it operates along the channel axis. The channel filter requires a preliminary step of transposing and flattening the original feature map, \textit{i.e.}, $H \times W \times c \rightarrow c \times (H \times W)$.

To expedite the transference of foundational knowledge acquired through extensive training on natural images, to the specialized domain of remote sensing imagery, we have integrated low-rank tunable parameters, \textit{i.e.}, LoRA\cite{hu2021lora}, within each Transformer layer within the SAM encoder. LoRA can preserve the model's original knowledge by maintaining the majority of the existing parameters in a frozen state. It is important to highlight that, to capture the dynamics of bi-temporal features efficiently, we have strategically incorporated the BCSF exclusively following the four global attention layers, leaving the other local window attention layers fixed. Moreover, given the distinct requirements for semantic level in change detection versus change captioning tasks, we have employed two-tiered features with different channel dimensions for task decoding. 
Specifically, we utilize features that have undergone a dimension reduction from $c$ to 256 for change detection, while retaining the original, unreduced features for captioning.

\subsection{Multi-task Semantic Aggregation Neck}

The encoder of the foundation model with substantial computational load is frequently frozen during the fine-tuning for downstream tasks, impeding the convergence of multi-task learning towards a unified latent space. Furthermore, the cross-task interaction of semantic information is crucial for multi-task learning. In this section, we introduce the proposed multi-task semantic aggregation neck to synthesize and integrate knowledge sharing across tasks. Specifically, the neck comprises $N$ multi-task semantic aggregation units, each equipped with intra-task attention to mine task-specific knowledge, and inter-task attention to amalgamate homogeneous information from different tasks.

\begin{figure}[!htbp]
    \centering
    \includegraphics[width=0.7\linewidth]{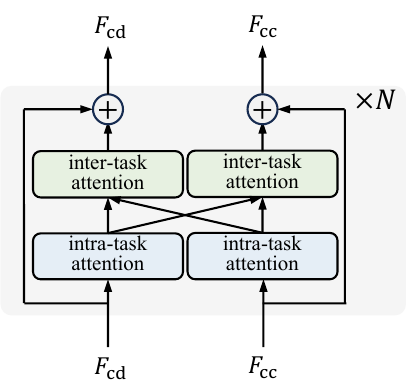}
    \caption{The overview of the multi-task semantic aggregation neck.}
    \label{fig:agg_neck}
\end{figure}

The structure of the unit is depicted in Fig. \ref{fig:agg_neck} and can be illustrated mathematically as follows,
\begin{align}
    \begin{split}
        F_{\text{cc}} &=  \Phi_{\text{cat}} (F_1, F_2) \\
        F_{\text{cc}}^{\text{intra}} &= \Phi_{\text{att}}^{\text{intra}} (F_{\text{cc}}) \\
        F_{\text{cc}} &= F_{\text{cc}} + \Phi_{\text{att}}^{\text{inter}} (F_{\text{cc}}^{\text{intra}}, F_{\text{cd}}^{\text{intra}}) \\
    \end{split}
\end{align}
where we take the change captioning as a case in point. The multi-task semantic aggregation unit processes the bi-temporal image features extracted by the encoder, and outputs inter-task attented features. Given that the aggregation is conducted across various tasks, we concatenate ($\Phi_{\text{cat}}$) the bi-temporal features along the batch dimension for further processing.
$\Phi_{\text{att}}^{\text{intra}}$ denotes intra-task attention, \textit{i.e.}, a multi-head self-attention Transformer layer. Conversely, $\Phi_{\text{att}}^{\text{inter}}$ represents inter-task attention, which serves to integrate change detection semantics into the change captioning task. Considering different channel dimensions between the features of the two tasks, we introduce an innovative cross-task attention mechanism based on bilinear similarity. The mechanism can effectively harness the homogeneous features from the heterogeneous semantic and pixel-level tasks, thereby significantly bolstering the efficacy of multi-task learning paradigms. 
\begin{figure}[!htbp]
    \centering
    \includegraphics[width=0.85\linewidth]{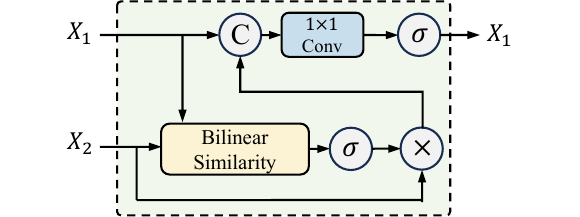}
    \caption{The structure of the inter-task attention unit.}
    \label{fig:inter_att}
\end{figure}

The inter-task attention structure, as depicted in Fig. \ref{fig:inter_att}, can be written by the following formula,
\begin{align}
    \begin{split}
        S &= X_1^T W X_2 \\
        \Delta X_2 &= X_2 \otimes \sigma (S) \\
        X_1 &= \sigma( \Phi_{\text{proj}} (\Phi_{\text{cat}} (X_1, \Delta X_2 ) ) ) \\
    \end{split}
\end{align}
where the equation illustrates the incorporation of features from task 2 ($X_2$) into task 1 ($X_1$), \textit{i.e.}, $\Phi_{\text{att}}^{\text{inter}}(X_1, X_2)$. $W$ symbolizes a matrix of learnable parameters, $S$ indicates the computed similarity, $\otimes$ is pixel-wise multiplication, $\sigma$ represents the sigmoid function, and $\Phi_{\text{cat}}$ signifies the concatenation performed along the channel dimension. $\Phi_{\text{proj}}$ projects the feature map to its original channel dimensions through a $1 \times 1$ convolution layer.

\subsection{Change Detection Decoder}

We have proposed a streamlined multi-scale change detection decoder for narrowing the gap between pixel-level interpretation and extracted dense semantic features. Specifically, We concatenate bi-temporal image features from the neck along the channel dimension and utilize SimpleFPN \cite{li2022exploring} to generate multi-scale feature maps. These maps are then resized into a unified scale to facilitate the decoding of the change mask, as delineated in the following formulation,
\begin{align}
    \begin{split}
        \{ F_i \} &= \Phi_{\text{sampling}} ( \Phi_{\text{cat}} ( F_{\text{cd}}^1, F_{\text{cd}}^2) ) \\
        F_i &= \Phi_{\text{resize}} (\Phi_{\text{conv}} (F_i) ) \\
        M &= \Phi_{\text{proj}} (\Phi_{\text{cat}} (\{ F_i \} ) ) \\
    \end{split}
\end{align}
where $F_{\text{cd}}^1$ and $F_{\text{cd}}^2$ denote the pre- and post-temporal semantic feature maps with the change detection tasks, respectively. $\Phi_{\text{sampling}}$ represents the multi-scale construction process based on the SimpleFPN \cite{li2022exploring}. The multi-scale feature maps $\{ F_i \in \mathbb{R}^{\frac{H}{2^{i+1}} \times \frac{W}{2^{i+1}} \times 512} \}, i \in \{1, 2, 3, 4 \}$ are subsequently refined to enhance their spatial characteristics via convolution. $\Phi_{\text{resize}}$ refers to bilinear interpolation to unify the spatial resolution, while $\Phi_{\text{proj}}$ projects the channel dimension of the feature maps to the class number using a $1 \times 1$ convolution layer.

\subsection{Change Caption Decoder}

We have developed a change captioning decoder based on the Vicuna LLM \cite{zheng2024judging}. Vicuna is fine-tuned based on LLaMA \cite{touvron2023llama}, delivering exceptional performance at a comparatively low cost. In contrast to the decoders used for vision language modeling in natural images, the change captioning task presents unique challenges due to the heterogeneity of the input (dual images) and the complexity of the semantics to be interpreted (effective changes).
To overcome these challenges, we have proposed a decoder with the capacity to perceive change semantics, which primarily comprises three components: a Q-Former \cite{li2023blip} that queries semantic features associated with different temporal phases as per human instructions; a change semantic feature enhancer that generates bi-temporal differential features; and a Vicuna LLM decoder that produces change descriptions. The process can be delineated by the following formula,
\begin{align}
    \begin{split}
        F_{\text{cc}}^{\{1,2\}} &= \Phi_{\text{q-former}} ( F_{\text{cc}}^{\{1,2\}}, P ) \\
        \Delta F_{\text{cc}} &= \Phi_{\text{enhancer}} ( F_{\text{cc}}^1, F_{\text{cc}}^2 ) \\
        T &= \Phi_{\text{llm-decoder}} ( \Phi_{\text{prompter}} (\Delta F_{\text{cc}}, P) ) \\
    \end{split}
\end{align}
where the bi-temporal feature maps undergo sequential processing by the Q-Former ($\Phi_{\text{q-former}}$) with shared parameters. $P$ denotes the task-specific instructions that are meticulously designed. $\Phi_{\text{enhancer}}$ is employed to generate change features with the $F_{\text{cc}}^1$ and $F_{\text{cc}}^2$, while $\Phi_{\text{prompter}}$ serves as a template operation that combines the produced differential features into the manual instructions. $\Phi_{\text{llm-decoder}}$ procures the final change description ($T$). Both the Q-Former and the LLM decoder strictly conform to the structure of MiniGPT-4 \cite{zhu2023minigpt} and inherit its pre-trained weights. 

\begin{figure}[!htbp]
    \centering
    \includegraphics[width=0.9\linewidth]{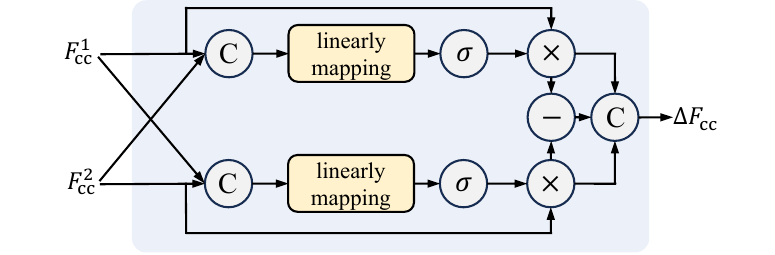}
    \caption{Change semantic feature enhancer}
    \label{fig:enhancer}
\end{figure}

The structure of the enhancer is illustrated in Fig. \ref{fig:enhancer} and can be described by the subsequent formula,
\begin{align}
    \begin{split}
        {F_{\text{cc}}^1}^{\prime} &= F_{\text{cc}}^1 \otimes \sigma( \Phi_{\text{proj-1}} (\Phi_{\text{cat}} (F_{\text{cc}}^1, F_{\text{cc}}^2) ) )  \\
        {F_{\text{cc}}^2}^{\prime} &= F_{\text{cc}}^2 \otimes \sigma( \Phi_{\text{proj-2}} (\Phi_{\text{cat}} (F_{\text{cc}}^2, F_{\text{cc}}^1) ) )  \\
        \Delta F_{\text{cc}} &= \Phi_{\text{cat}} ( {F_{\text{cc}}^1}^{\prime}, {F_{\text{cc}}^2}^{\prime}, {F_{\text{cc}}^1}^{\prime} - {F_{\text{cc}}^2}^{\prime} ) \\
        \Delta F_{\text{cc}} &= \Phi_{\text{proj}} ( \Delta F_{\text{cc}} ) \\
    \end{split} \label{eq:enhancer}
\end{align}
where $\Phi_{\text{cat}}$ symbolizes the vector concatenation along the channel dimension. $\Phi_{\text{proj-1}}$ and $\Phi_{\text{proj-2}}$ denote linearly mapping the features to a single channel number. $\Phi_{\text{proj}}$ projects the dimensions of the enhanced feature to that of the LLM's hidden state. To reconcile the semantic gap between change descriptions and natural dialogue scenarios, we have integrated LoRA into the decoder of the LLM.

\subsection{Training Strategy} \label{sec:training_strategy}

Imagine we are given a training set $\mathcal{D}_{\text{train}} = \{(\mathcal{I}_1, y_1), \dots, (\mathcal{I}_N, y_N) \}$, where $\mathcal{I}_i = \{\mathcal{I}_i^1 \in \mathbb{R}^{H \times W \times 3}, \mathcal{I}_i^2 \in \mathbb{R}^{H \times W \times 3} \} $ represents the bi-temporal remote sensing images, and $y_i = \{y_i^\text{cd} \in \mathbb{R}^{H \times W}, y_i^\text{cc} \in \mathcal{C}^{N} \}$ denotes corresponding change detection and change description annotations, respectively. 
However, in practical settings, we frequently encounter a dearth of annotations that concurrently encompass both $y_{\text{cd}}$ and $y_{\text{cc}}$. We omit the subscript $i$ for the sake of simplicity. 
This is exemplified in datasets such as LEVIR-CD \cite{chen2020spatial} and LEVIR-CC \cite{liu2022remote}, where merely a subset of image pairs incorporate both labels. We designate the training data that includes CD, CC, and CD $\&$ CC annotations as $\mathcal{D}_{\text{train}}^{\text{cd}}$, $\mathcal{D}_{\text{train}}^{\text{cc}}$, and $\mathcal{D}_{\text{train}}^{\text{cd-cc}}$, correspondingly. In addition, we establish the loss function for multi-task learning as,
\begin{align}
    \begin{split}
        \mathcal{L}_{\text{cd}} &= -\frac{1}{N}\sum_{i}^{N} ( y_i \log (\hat{y}_i) + (1-y_i)  \log (1 - \hat{y}_i) ) \\
        \mathcal{L}_{\text{cc}} &= -\frac{1}{M}\sum_{j}^{M} w_j \log (\hat{w}_j) \\
        \mathcal{L} & = \mathcal{L}_{\text{cc}} + \lambda \mathcal{L}_{\text{cd}} \\
    \end{split}
\end{align}
where $\hat{y}_i$ and $y_i$ symbolize the prediction and annotation of an individual pixel in change detection, respectively. The total number of pixels is represented by $N$. The one-hot encodings and the predicted caption of the $j$th word are denoted by $w_j$ and $\hat{w}_j$, respectively. $M$ is the total number of caption tokens. $\lambda$ is utilized to maintain a balance in the loss.

To facilitate a smooth progression of the training process, we have formulated the subsequent training strategy: Throughout each epoch, 
i), training CD with $\mathcal{D}_{\text{train}}^{\text{cd}}$; 
ii), training CC with $\mathcal{D}_{\text{train}}^{\text{cc}}$, keeping image encoder frozen; 
iii), joint training neck with $\mathcal{D}_{\text{train}}^{\text{cd-cc}}$, keeping other parameters frozen.

\section{Experimental Results and Analyses}
\subsection{Datasets}

This paper investigates remote sensing image change captioning underpinned by the latent knowledge of the foundation model and the semantics of change detection. During the experiments, the LEVIR series datasets were leveraged for both training and evaluation purposes \cite{chen2020spatial, liu2022remote}. These datasets encompass the change detection dataset, LEVIR-CD \cite{chen2020spatial}, and the change captioning dataset, LEVIR-CC \cite{liu2022remote}. The following provides an in-depth introduction to these datasets.

\textbf{LEVIR-CD} \cite{chen2020spatial}: The LEVIR-CD dataset is comprised of 637 pairs of high-resolution (VHR, 0.5m/pixel) images obtained from Google Earth, each measuring $1024 \times 1024$ pixels. These bi-temporal images span a period ranging from 5 to 14 years, thereby capturing significant changes in land use, particularly in relation to the construction and demolition of buildings. The fully annotated LEVIR-CD dataset encompasses a total of 31,333 instances of building changes, and we adhere to the official division of training and test sets.

\textbf{LEVIR-CC} \cite{liu2022remote}: This dataset derives its images from LEVIR-CD. The LEVIR-CC dataset modifies the bi-temporal images to a size of 256×256 pixels, resulting in a total of 10,077 image pairs. For each image pair, five distinct annotations were collected from five different annotators to delineate the differences between the images, culminating in a total of 50,385 annotations. For our purposes, we utilized 7,590 pairs for the training set, 1,438 pairs for the validation set, and 2,135 pairs for the test set.

\subsection{Evaluation Metrics}

For the task of change captioning, we employ widely used image captioning metrics such as BLEU \cite{papineni2002bleu}, METEOR \cite{banerjeemeteor}, ROUGE$_L$ \cite{lin2004rouge}, and CIDERr \cite{vedantam2015cider} for evaluation. The scores from these metrics are directly proportional to the similarity between the generated sentences and the reference sentences, thereby serving as an indicator of the model's performance. For the change detection, our focus lies on the Precision (P), Recall (R), F1-score (F1), Intersection over Union (IoU) about the change category, and the overall segmentation accuracy (OA).

\subsection{Implementation Details}

The model presented in this study is implemented using Pytorch on the NVIDIA A100 platform. The ViT encoder of the large version SAM is utilized as the bi-temporal image feature extractor, while the Vicuna-7B model serves as the large language decoder for change captioning. The weights for the image encoder and caption decoder are initialized from the SAM and Mini-GPT4 models, respectively. To address domain differences during training, LoRA efficient parameter fine-tuning is applied to the image encoder and caption decoder, with specific parameters including rank $=$ 16, LoRA alpha $=$ 32, LoRA dropout $=$ 0.05, and no bias. The remaining parameters of the encoder, caption decoder, and Q-Former are kept frozen. The AdamW optimizer with an initial learning rate of $1e-4$ is employed for training, along with a cosine annealing scheduler with a linear warm-up to adjust the learning rate. The training process consists of 300 epochs with a batch size of 1, and the token with the highest probability is selected for output during caption generation. The number of neck units is set to 3, and the balance factor in the loss function is $\lambda = 0.5$. Various instruction prompts are used during training, such as ``Describe the difference between the new remote sensing image and the old one in detail.", ``What is the main change between the two remote sensing scenes? Describe it in detail.", ``Please provide a detailed description of the difference between these two remote sensing pictures.", and ``Can you describe what has been changed between these two remote sensing pictures for me?". During testing, a multi-output ensemble approach is utilized to generate the final caption.

\begin{figure*}[!htbp]
    \centering
    \includegraphics[width=0.99\linewidth]{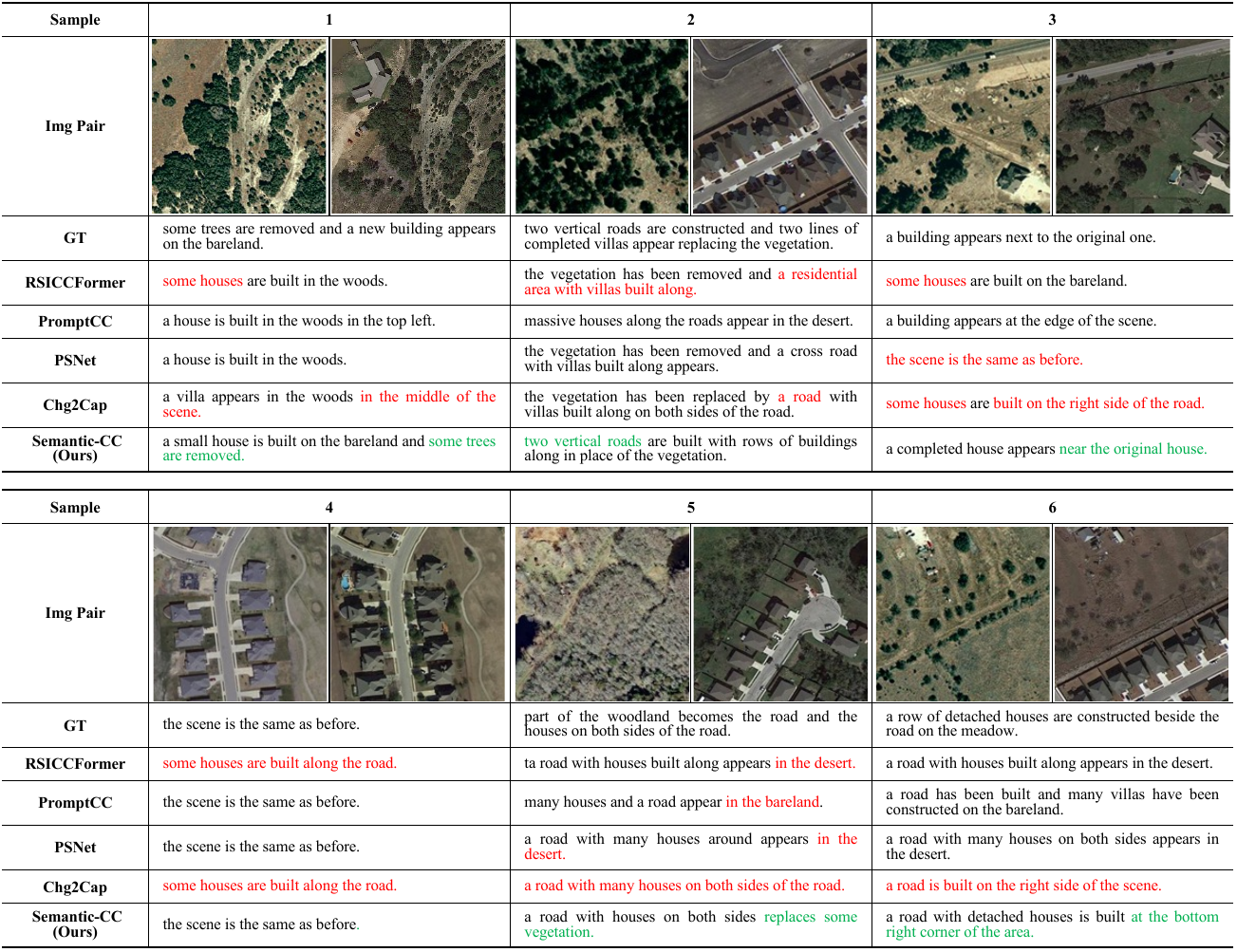}
    \caption{Qualitative comparison results with other methods on the LEVIR-CC dataset, where red indicates incorrect outputs and green indicates correct outputs.
    }
    \label{fig:comparison_sota}
\end{figure*}

\begin{table*}[!htbp]
  \centering
  \caption{Comparison with other state-of-the-art methods on LEVIR-CC test set.}
  \resizebox{0.8\linewidth}{!}{
    \begin{tabular}{c|ccccccc}
    \toprule
    Method   & METEOR & ROUGE$_L$ & CIDEr & BLEU-1 & BLEU-2 & BLEU-3 & BLEU-4 \\
    \midrule
    RSICCFormer \cite{liu2022remote} & 0.3961 & 0.7412 & 1.3412 & 0.8472 & 0.7627 & 0.6887 & 0.6277 \\
    PromptCC \cite{liu2023decoupling} & 0.3882 & 0.7372 & 1.3644 & 0.8366 & 0.7573 & 0.691 & 0.6354 \\
    PSNet \cite{liu2023progressive} & 0.3880 & 0.7360 & 1.3262 & 0.8386 & 0.7513 & 0.6789 & 0.6211 \\
    Chg2Cap \cite{chang2023changes} & 0.4003 & 0.7512 & 1.3661 & 0.8614 & 0.7808 & 0.7066 & 0.6439 \\
    \midrule
    Semantic-CC (ours) & \textbf{0.4058} & \textbf{0.7776} & \textbf{1.3851} & \textbf{0.8807} & \textbf{0.7968} & \textbf{0.7147} & \textbf{0.6451} \\
    \bottomrule
    \end{tabular}
    }
  \label{tab:comparison}
\end{table*}

\subsection{Comparison with the State-of-the-Art}

We compare the proposed Semantic-CC with a variety of state-of-the-art (SOTA) remote sensing image change captioning methods, including RSICCFormer \cite{liu2022remote}, PromptCC \cite{liu2023decoupling}, PSNet \cite{liu2023progressive}, and Chg2Cap \cite{chang2023changes}. These methods span architectures such as CNN, Transformer, and foundation models, as well as paradigms of both single-task and multi-task learning. The metrics for the comparative results are presented in Tab. \ref{tab:comparison}. It can be seen that Semantic-CC surpasses all SOTA methods in all metrics. The superiority is particularly noticeable in ROUGE$_L$, CIDEr, BLEU-1, BLEU-2, and BLEU-3. ROUGE$_L$, reflecting the recall, suggests that Semantic-CC can offer a more exhaustive description of scene changes, encapsulating the semantic descriptions present in the annotations. CIDEr and lower-order BLEU metrics gauge the precision, indicating that Semantic-CC can accurately delineate key changes. Overall, Semantic-CC, based on foundational knowledge and semantic guidance, is capable of delivering comprehensive and precise change captions.

To provide a more intuitive representation of the effects of different methods on change sentence generation, we offer qualitative visualizations in Fig. \ref{fig:comparison_sota}. Six representative images were selected for this evaluation. The generated results reveal that: i) Owing to the pixel-level semantic guidance, the proposed method can output more precise locations and quantities. For example, in sub-figures 1 and 2, RSICCFormer and Chg2Cap inaccurately describe the quantity and location, whereas Semantic-CC adeptly manages these complex scenarios and accurately outputs location and quantity semantics; ii) Leveraging the prior knowledge from foundation models, Semantic-CC can consistently handle complex scenarios and disturbances from illumination. For instance, in sub-figure 3, PSNet is unable to detect changes, in sub-figure 4, RSICCFormer and Chg2Cap erroneously output changes, and in sub-figure 5, other methods are unable to identify the semantic category of the forest, but Semantic-CC can still make the accurate prediction; iii) Semantic-CC can generate more standardized, fluent sentences that comply with human language systems. This is attributed to our proposed method's use of a large language model decoder, which can output change descriptions that align with human language conventions across a variety of scenarios.

\begin{figure*}[!htbp] 
    \centering
    \includegraphics[width=0.99\linewidth]{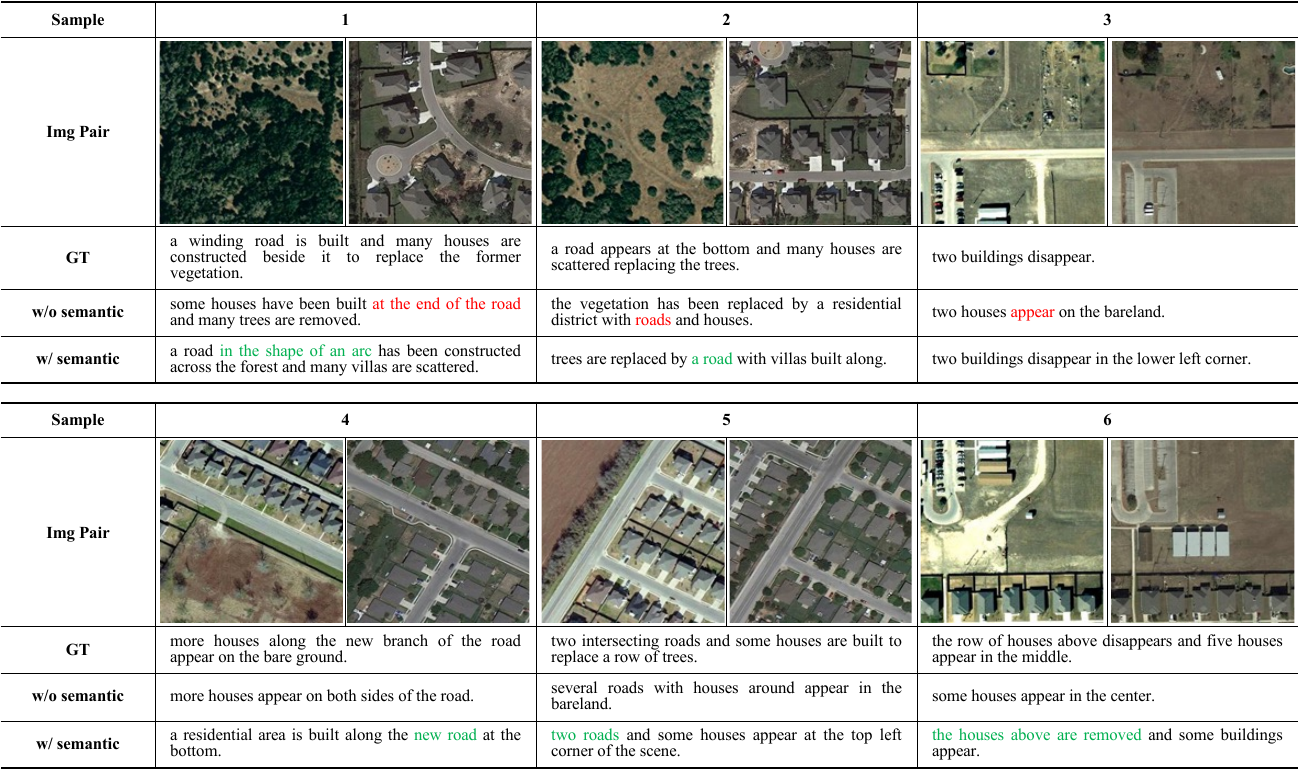}
    \caption{The impact of pixel-level change detection semantic guidance on change caption generation. Red indicates incorrect entity descriptions, while green indicates correct entity descriptions.}
    \label{fig:vis_tasks}
\end{figure*}

\subsection{Ablation Studies}

\subsubsection{Effects of Different Architecture Components}

In this section, we assess the effects of the proposed components on the performance of change captioning. We initially evaluate the impact of the bi-temporal change semantic filter (BCSF) and change semantic feature enhancer (enhancer) solely on the generation of change sentences within the change captioning task. Subsequently, we incorporate the multi-task semantic aggregation neck to broaden the scope to a multi-task learning framework, utilizing the neck of a full Transformer architecture for comparison, \textit{i.e.}, intra-task attention and inter-task attention corresponding to the Transformer self-attention and cross-attention respectively. The experimental results are presented in Tab. \ref{tab:ablation_components}.

\begin{table*}[!htbp] 
\centering
\caption{
Performance of different architecture components on LEVIR-CC test set. BCSF: bi-temporal change semantic filter. Enhancer: change semantic feature enhancer. Neck$_\text{cs}$ and Neck$_\text{sm}$ indicate taking the Transformer cross-attention mechanism or the bilinear similarity mechanism to form inter-task attention in the neck.
}
\label{tab:ablation_components}
\resizebox{0.8\linewidth}{!}{
\begin{tabular}{c c|c c|ccccccc}
    \toprule
    BCSF & Enhancer & Neck$_\text{cs}$ & Neck$_\text{sm}$ & METEOR & ROUGE$_L$ & CIDEr & BLEU-1 & BLEU-2 & BLEU-3 & BLEU-4 \\
    \midrule
    \ding{55} & \ding{55} & \ding{55} & \ding{55} & 0.3315 & 0.5806 & 1.2254 & 0.76436 & 0.6518 & 0.5538 & 0.4657 \\
    \checkmark & \ding{55} & \ding{55} & \ding{55} & 0.3573 & 0.6042 & 1.2992 & 0.8264 & 0.7072 & 0.6062 & 0.5275 \\
    \checkmark & \checkmark & \ding{55} & \ding{55} & 0.3821 & 0.7133 & 1.3427 & 0.8584 & 0.7545 & 0.6228 & 0.5742 \\
    \midrule
    \checkmark & \checkmark & \checkmark & \ding{55} & 0.3945 & 0.7512 & 1.3742 & 0.8701 & 0.7742 & 0.6921 & 0.6321 \\
    \checkmark & \checkmark & \ding{55} & \checkmark & \textbf{0.4058} & \textbf{0.7776} & \textbf{1.3851} & \textbf{0.8807} & \textbf{0.7968} & \textbf{0.7147} & \textbf{0.6451} \\
    \bottomrule
    \end{tabular}
}
\end{table*}

From table \ref{tab:ablation_components}, we can deduce the following: i) In single-task learning, both the BCSF and the enhancer contribute to performance improvements, particularly on the ROUGE$_L$ metric, which gauges the model's capacity to concentrate on effective changes. The BCSF and the enhancer can extract the effective changes of bitemporal images to enhance the model's performance. ii) The multi-task learning framework outperforms single-task learning across all metrics, suggesting the complementary nature of employing the pixel-level semantic guidance of the change detection task to generate change captions. iii) The application of the cross-task attention mechanism, based on bilinear similarity, is superior to the direct use of the cross-attention mechanism in the Transformer, especially on ROUGE$_L$, BLEU-2, BLEU-3, and BLEU-4, which indicates that the model's improvement is centered on a comprehensive perception of changes.

\begin{figure*}[!htbp] 
    \centering
    \includegraphics[width=0.9\linewidth]{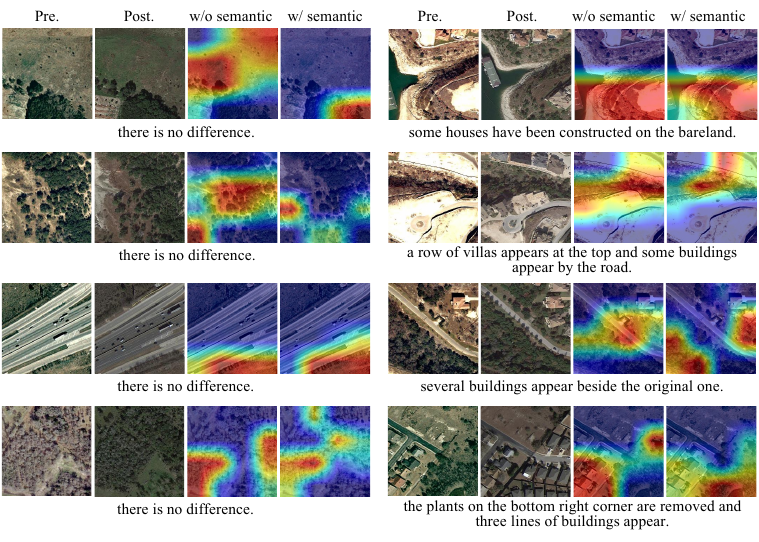}
    \caption{The heatmap of the bi-temporal differential features of pixel-level change detection semantic guidance, where darker colors indicate higher feature response values.}
    \label{fig:vis_tasks_heatmap}
\end{figure*}

To further scrutinize the advantages of pixel-level change detection semantic guidance for change caption generation, we present some description generation results in Fig. \ref{fig:vis_tasks}, and the spatial response heatmaps of the differential feature map in Fig. \ref{fig:vis_tasks_heatmap}. Fig. \ref{fig:vis_tasks} reveals that by incorporating a semantic guidance branch, \textit{i.e.}, the multi-task learning, the model can accurately delineate the location and quantity of changes. For instance, in sub-figures 1, 2, and 3, the model without semantic guidance inaccurately describes the location of changes, the number of roads, the presence or absence of changes, \textit{etc.}, whereas the model with semantic guidance can correctly perceive these semantics. Moreover, it can also better comprehend the scene, for example, in sub-figure 4, the model with semantic guidance can identify the appearance of new roads; in sub-figure 5, it can perceive the change in quantity and provide the location information of the change; in sub-figure 6, it can describe the increase or decrease of different building instances simultaneously.

Given that the features within the CC decoder have been deeply integrated with the text instructions, which are high-level semantic features, we visualized the bi-temporal differential features post-neck with heatmaps. From Fig. \ref{fig:vis_tasks_heatmap}, it is evident that i) for scenes without changes, \textit{i.e.}, the left sub-figures, the model with semantic guidance can generate lower and smaller differential feature responses; ii) for scenes with changes, \textit{i.e.}, the right sub-figures, the model can generate more focused and accurate responses, which are essentially consistent with the areas where changes have occurred.

\subsubsection{Effects of the Structure of Change Semantic Feature Enhancer in CC Decoder}

Given the significant differences in input and attented semantics compared to single-image visual-language models, we developed a change semantic feature enhancer for change description within the decoder. This enhancer generates bi-temporal differentiated features, enabling the LLM to produce change descriptions, as depicted in Eq. \ref{eq:enhancer}. We evaluated the effects of the enhancer's position (before or after the Q-Former) and the enhancer's structure on the performance of the change description. For the sake of simplicity, we carried out experiments on the CC task devoid of semantic guidance. We define the first two formulas for temporal feature interaction in Eq. \ref{eq:enhancer} as activation operations, and the subtraction of bi-temporal features as differentiation operations. These two operations are amalgamated to form various structures of the enhancer.

\begin{table*}[!htbp]
  \centering
  \caption{Performance of different positions and structures of the change semantic feature enhancer in captioning decoder on LEVIR-CC test set. Pre./Post.: the enhancer before/after the Q-Former. Act.: temporal feature interaction. Sub.: bi-temporal feature subtraction.}
  \resizebox{0.7\linewidth}{!}{
    \begin{tabular}{cc|cc|ccccccc}
    \toprule
    Pre. & Post. & Act. & Sub. & METEOR & ROUGE$_L$ & CIDEr & BLEU-1 & BLEU-2 & BLEU-3 & BLEU-4 \\
    \midrule
    \checkmark & \ding{55} & \ding{55} & \checkmark & 0.3342 & 0.5914 & 1.2143 & 0.7941 & 0.6991 & 0.5545 & 0.4551 \\
    \ding{55} & \checkmark & \ding{55} & \checkmark & 0.3483 & 0.6042 & 1.2594 & 0.8264 & 0.7072 & 0.6062 & 0.5275 \\
    \midrule
    \checkmark & \ding{55} & \checkmark & \ding{55} & 0.3546 & 0.6547 & 1.2794 & 0.8184 & 0.7181 & 0.6174 & 0.5432 \\
    \ding{55} & \checkmark & \checkmark & \ding{55} & 0.3714 & 0.7043 & 1.3053 & 0.8438 & 0.7274 & 0.6221 & 0.5621 \\
    \midrule
    \checkmark & \ding{55} & \checkmark & \checkmark & 0.3696 & 0.6954 & 1.2977 & 0.8384 & 0.7143 & 0.6164 & 0.5591 \\
    \ding{55} & \checkmark & \checkmark & \checkmark  & \textbf{0.3821} & \textbf{0.7133} & \textbf{1.3427} & \textbf{0.8584} & \textbf{0.7545} & \textbf{0.6228} & \textbf{0.5742} \\
    \bottomrule
    \end{tabular}
    }
  \label{tab:ablation_enhancer_structure}%
\end{table*}

\begin{table*}[!htbp]
  \centering
  \caption{Performance of different training strategies on LEVIR-CC test set.}
  \resizebox{0.7\linewidth}{!}{
    \begin{tabular}{c|ccccccc}
    \toprule
    Strategy & METEOR & ROUGE$_L$ & CIDEr & BLEU-1 & BLEU-2 & BLEU-3 & BLEU-4 \\
    \midrule
    1-stage & 0.3501 & 0.7335 & 1.3053 & 0.8051 & 0.7017 & 0.6021 & 0.5622 \\
    2-stage & 0.3895 & 0.7402 & 1.3591 & 0.8691 & 0.7737 & 0.6834 & 0.6223 \\
    3-stage & \textbf{0.4058} & \textbf{0.7776} & \textbf{1.3851} & \textbf{0.8807} & \textbf{0.7968} & \textbf{0.7147} & \textbf{0.6451} \\
    \bottomrule
    \end{tabular}
    }
  \label{tab:training_strategy}
\end{table*}

\begin{figure*}[!htbp]
    \centering
    \includegraphics[width=0.9\linewidth]{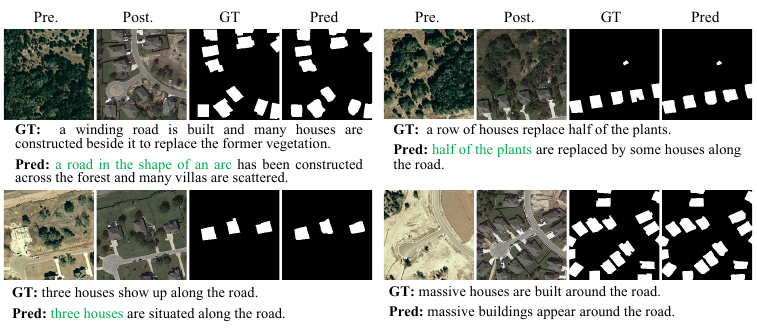}
    \caption{Visualization results of change detection and change caption on the LEVIR-CC dataset.}
    \label{fig:vis_cd}
\end{figure*}

The experimental results are presented in Tab. \ref{tab:ablation_enhancer_structure}. The table leads us to the following observations: i) Positioning the enhancer after the Q-Former results in optimal performance, as the pre-trained Q-Former, designed to model a single image, encounters difficulties when processing differentiated features. ii) Employing solely the differentiation operation can generate differentiated features, but it fails to model the effective changes of bi-temporal features and to filter out irrelevant information. Utilizing only the activation operation can model the shared semantics of bi-temporal features, but it cannot directly capture differentiated features. Both of these structures lead to a certain degree of performance degradation. However, their integration within the enhancer can ignore interference and grasp changes simultaneously, to expedite network convergence and yield superior performance.

\subsubsection{Effects of Different Training Strategies}

To ensure a stable convergence in multi-task learning and enable pixel-level semantic change detection to guide the enhancement of change captioning, we conducted experiments with various training strategies. These strategies are categorized based on the frequency of dataset switches per epoch into single-stage, two-stage, and three-stage. The three-stage strategy, as detailed in Sec. \ref{sec:training_strategy}, was ultimately adopted in this study. The two-stage strategy involves initial training of CD, followed by CC, while the single-stage strategy entails a combined training of CD and CC. The experimental results, as presented in Tab. \ref{tab:training_strategy}, indicate that the three-stage training yields the most optimal performance in the change captioning task. This can be attributed to the model's initial optimization in accordance with individual objectives during the training process, followed by a joint optimization. This strategy not only alleviates tasks' coupling but also circumvents the risk into a local optimum. Furthermore, it facilitates a faster convergence speed for the target task.

\subsubsection{Comparison with the State-of-the-Art on CD}

\begin{table}[!tbp]
  \centering
  \caption{Comparison with other state-of-the-art CD methods on LEVIR-CD test set.}
    \resizebox{0.9\linewidth}{!}{
    \begin{tabular}{c|ccccc}
    \toprule
    Method & P     & R     & F1    & IoU   & OA \\
    \midrule
    IDET \cite{guo2022idet}  & 91.3  & 86.6  & 88.9  & 80.2  & 98.1 \\
    ChangeFormer \cite{bandara2022transformer} & 92.1  & 88.8  & 90.4  & 82.5  & 99.0 \\
    WNet \cite{tang2023wnet}  & 91.2  & 90.2  & 90.7  & 83.0    & 99.1 \\
     CSTSUNet \cite{wu2023cstsunet} & 92.0   & 89.4  & 90.7  & 83.0    & 99.1 \\
    TTP \cite{chen2023time}   & 93.0    & 91.7  & 92.1  & 85.6  & 99.2 \\
    Semantic-CC (Ours) & \textbf{93.0} & \textbf{91.8} & \textbf{92.4} & \textbf{85.8} & \textbf{99.2} \\
    \bottomrule
    \end{tabular}
    }
  \label{tab:comparison_cd}
\end{table}

In this section, we compare the output of the CC branch with other state-of-the-art CD methods, including ChangeFormer \cite{bandara2022transformer}, IDET \cite{guo2022idet}, WNet \cite{tang2023wnet}, CSTSUNet \cite{wu2023cstsunet}, and TTP \cite{chen2023time}. Notably, TTP is a CD algorithm developed based on the SAM foundation model, currently demonstrating the most superior performance on the LEVIR-CD dataset. The comparative results, as presented in Tab. \ref{tab:comparison_cd}, reveal that Semantic-CC also exhibits outstanding performance in the CD task, achieving metrics comparable to TTP. Importantly, it does not induce negative transfer due to multi-task learning, \textit{i.e.}, the learning of one task does not compromise the performance of another. Additionally, we provide several visual segmentation examples in Fig. \ref{fig:vis_cd}, illustrating that Semantic-CC is capable of generating high-precision CD masks and CC descriptions. It is noteworthy that in the dataset utilized in this study, CD only provides segmentation masks for building changes, and the semantic vocabulary of CC is more extensive than that of CD. Our strategy of leveraging pixel-level semantics from CD to guide the training of the CC task is rational, and the disparity in semantic vocabulary does not adversely affect their respective performances.

\section{Conclusion}

This paper introduces Semantic-CC, a novel method for remote-sensing image change captioning, based on foundational knowledge and semantic guidance. By leveraging the latent knowledge inherent in foundation models, Semantic-CC enhances the model's generalization capabilities across diverse spatio-temporal scenarios. Additionally, it offers pixel-level semantic guidance by incorporating a change detection task, thereby yielding more granular and precise descriptions. Semantic-CC is composed of four key components: a bi-temporal SAM-based encoder for the extraction of dual-image features where a change semantic filter for bi-temporal interaction; a multi-task semantic aggregation neck that utilizes intra- and inter-task information attention for dual-task interaction; a change detection decoder that provides pixel-level semantic guidance for change captioning; and a Vicuna-based change caption decoder for the generation of change description sentences.
To ensure the smooth convergence of CD and CC, and to prevent negative transfer in multi-task learning, we have also devised a three-stage training strategy. Experimental results on the LEVIR-CD and LEVIR-CC datasets substantiate the complementary nature of the semantics of CC and CD. They also illustrate that Semantic-CC is capable of producing high-precision, granular CD masks and CC descriptions.

\ifCLASSOPTIONcaptionsoff
  \newpage
\fi



\bibliographystyle{IEEEtran}
\small{
\bibliography{IEEEabrv,myreferences}
}

\end{document}